\documentclass[10pt,twocolumn,letterpaper]{article}

\usepackage{cvpr}
\usepackage{times}
\usepackage{epsfig}
\usepackage{graphicx}
\usepackage{amsmath}
\usepackage{amssymb}

\usepackage{booktabs}
\usepackage{array}
\usepackage{subcaption}
\usepackage{cite}
\usepackage{color}

\usepackage[breaklinks=true,bookmarks=false]{hyperref}

\cvprfinalcopy 

\def\cvprPaperID{5248} 

\begin{document}

\title{Weakly Supervised Deep Image Hashing through Tag Embeddings}

\author{Vijetha Gattupalli, Yaoxin Zhuo, Baoxin Li\\
Arizona State University\\
{\tt\small \{vijetha.gattupalli, yzhuo6, baoxin.li\}@asu.edu}
}


\maketitle

\begin{abstract}
   Many approaches to semantic image hashing have been formulated as supervised learning problems that utilize images and label information to learn the binary hash codes. However, large-scale labelled image data is expensive to obtain, thus imposing a restriction on the usage of such algorithms. On the other hand, unlabelled image data is abundant due to the existence of many Web image repositories. Such Web images may often come with images tags that contains useful information, although raw tags in general do not readily lead to semantic labels. Motivated by this scenario, we formulate the problem of semantic image hashing as a weakly-supervised learning problem. We utilize the information contained in the user-generated tags associated with the images to learn the hash codes. More specifically, we extract the word2vec semantic embeddings of the tags and use the information contained in them for constraining the learning. Accordingly, we name our model Weakly Supervised Deep Hashing using Tag Embeddings (WDHT). WDHT is tested for the task of semantic image retrieval and is compared against several state-of-art models. Results show that our approach sets a new state-of-art in the area of weekly supervised image hashing.
\end{abstract}

\section{Introduction}
\begin{table}
\begin{center}
\begin{tabular}{|>{\centering\arraybackslash}m{0.5cm}|>{\centering\arraybackslash}m{2.2cm}>{\centering\arraybackslash}m{2.2cm}>{\centering\arraybackslash}m{2.2cm}|}
         \hline
         & Sample 1 & Sample 2 & Sample 3  \\
         \hline
        \rotatebox{90}{\textbf{Images}} &
        \raisebox{-\totalheight}{\includegraphics[width=2.2cm, height=2.2cm]{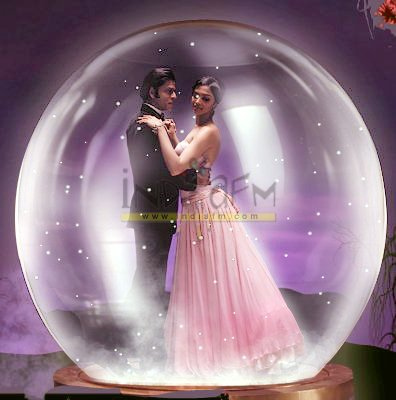}} &  \raisebox{-\totalheight}{\includegraphics[width=2.2cm, height=2.2cm]{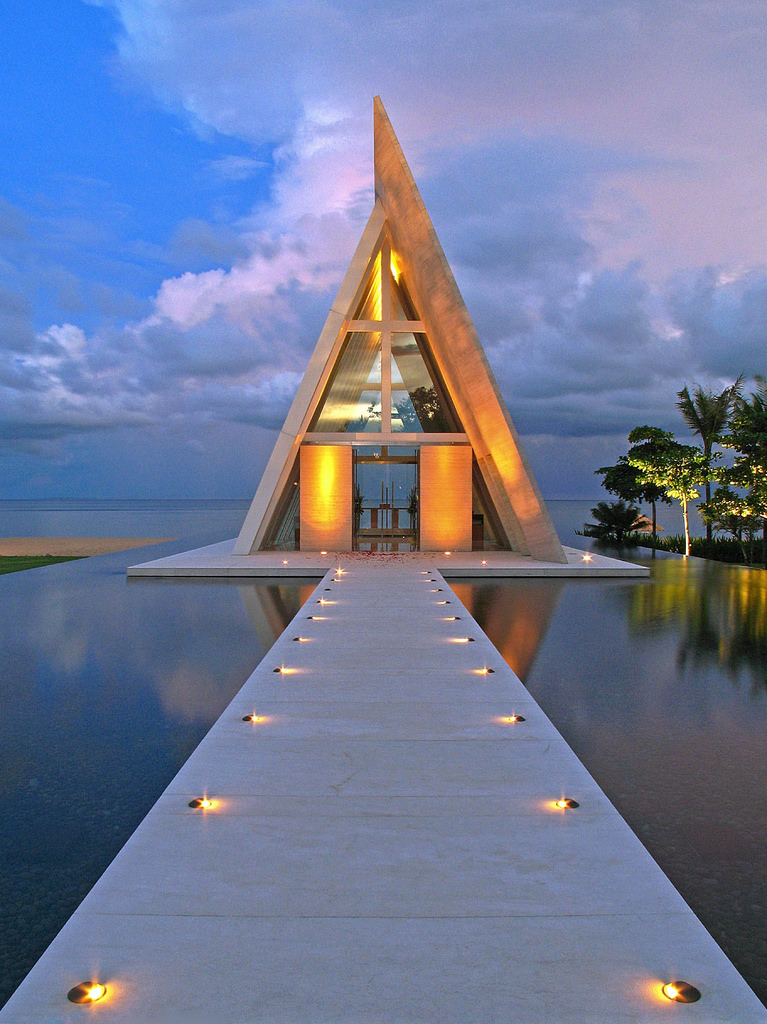}}&
        \raisebox{-\totalheight}{\includegraphics[width=2.2cm, height=2.2cm]{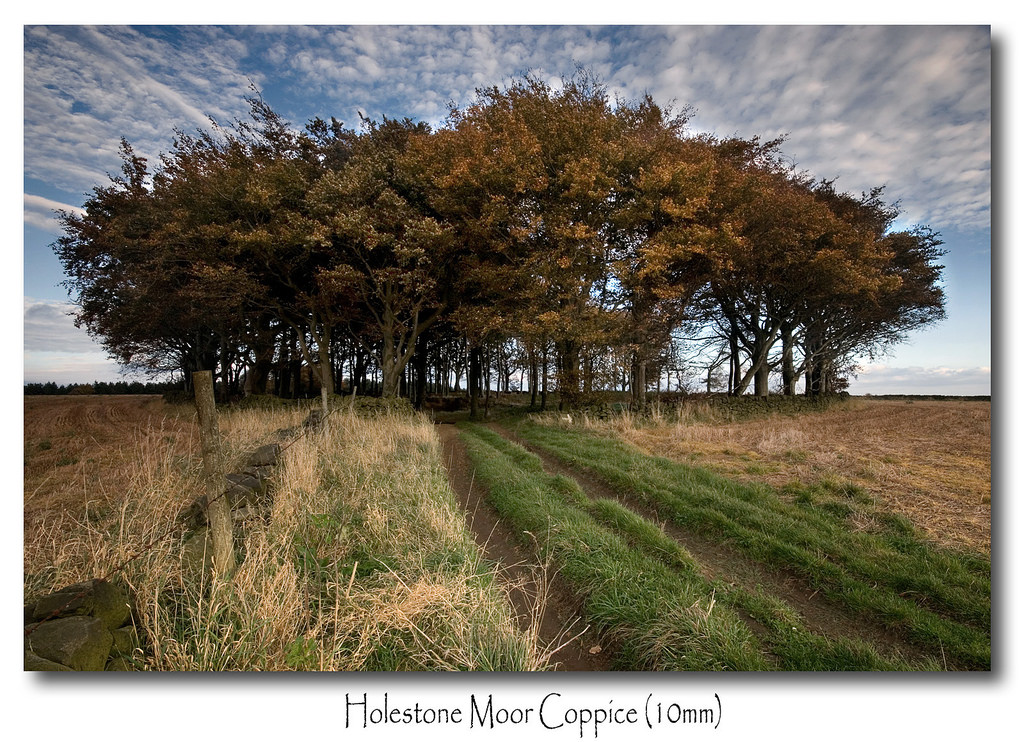}}\\
        \rotatebox{90}{\textbf{Tags}} &
        \tiny {\textit{\#india \#cinema \#movie \#star \#still \#handsome \#bollywood \#actor \#khan \#shahrukh \#srk \#omshantiom}} & 
        \tiny{\textit{\#sunset \#bali \#reflections \#indonesia \#mirror \#asia \#boda \#mariage \#hochzeit \#indonesien \#heirat \#chappel \#conradhotel \#50faves \#justimagine \#weddingchappel \#perfectangle \#infinestyle \#megashot \#theroadtoheaven \#thegoldendreams}} & 
        \tiny{\textit{ \#wood \#trees \#fence \#track \#derbyshire \#farming \#wideangle \#fields \#agriculture \#grassland \#sigma1020 \#autums \#marlock \#holestone \#holestonemoor}} \\ 
        \rotatebox{90}{\textbf{Labels}} & \scriptsize {dancing} & \scriptsize {buildings, clouds, reflection, sky, sunset} & \scriptsize{grass, sky,     tree} \\ 
         \hline
\end{tabular}
\end{center}
\caption{Table showing the image-tag-label triplets for some random samples from NUS-WIDE dataset.} \label{tab:cap}
\end{table}

Semantic Image Hashing has been an active research area for the past few years due to its ability to both search and store massive image databases efficiently. Briefly, it is the task of semantically mapping images to binary codes such that some notion of similarity is preserved. In this setting, similarity is determined by the ground truth class labels, which are expensive to obtain. This imposes a restriction on the amount of training data available. On the other hand,  much web image data available today have associated textual meta-data (tags). Such tag information is often readily available and is inexpensive. Owing to these facts, in this paper, we attempt the problem of  weekly supervised semantic image hashing by leveraging the tag information associated with the Web images. 

The current problem is addressed as weakly supervised mainly due to the following reasons. Tags may contain some information related to the semantics of the images. However, it is non-trivial to extract explicit label information from raw tags. Table 1 illustrates three samples from the NUS-WIDE dataset. It can be noticed that sample1 has no tags that are directly associated with the label ``dancing". While samples 2 and 3 have some tags that convey label information, they still have their own shortcomings. For example, they are associated with too many uninformative tags.These uninformative tags may be a consequence of the social-media behaviour of the public like opinion expression, self presentation, attracting attention etc \cite{gupta2010survey}. This results in tags that may be subjective (eg. \textit{\#thegoldendreams, \#handsome, \#50faves}), purely context oriented (eg. \textit{\#india, \#conradhotel \#katrina}), photography related (\textit{\#wideangle}) etc. Thus these tags contain information which is not related to the image content, making the process of extracting labels from tags further difficult. There are some prior works \cite{gupta2010survey}, \cite{sen2007quest} that attempted to address the difficulties in extracting information from raw tags. 



Even though our work focuses on using tag information to assist in learning the hash space, our algorithm does not fall under the category of cross-modal hashing (CMH). CMH deals with learning hash spaces that are shared for samples from various modalities. Ideally, a space thus learnt should be able to retrieve samples from one modality by using query samples from a different modality (e.g., retrieving images/videos using text queries and vice versa) \cite{wang2016comprehensive}. Our work only deals with direct image hashing where the query and retrieval samples are images. We only utilize the information from tags to learn better hash spaces for semantic image retrieval. Further, much work in CMH assumes the availability of image-tag-label triplets and use this information to learn the shared hash space. Thus they can be called supervised learning approaches, while ours is a weakly supervised approach. 

A key component of our method is the utilization of the \textit{word2vec} model \cite{mikolov2013efficient}, which is a method for embedding English words onto a vector space such that the cosine similarity between the vectors of the words is in accordance with their semantic similarity. In our task, the $<$image,tag set$>$ pairs are from the Web image datasets, and the tags generally bear some relevance to the semantics of the image (albeit this relevance may be weak, noisy, and incomplete). Hence we employ the \textit{word2vec} representation of the tags in our model, and regularize the learned hash space in such a way that images having similar tag vectors should have similar hash codes. Using the word vectors of the tags may lead to a better semantic hash space as compared to using only the binary tag vectors themselves. For example, if the training data contains images of cats and dogs, and several other non-animal classes, we would want the hash sub-spaces of the cats and the dogs to be close to each other. Further, an animal in a test set (for example horse), whose true class is not defined in the training set would ideally be mapped to a code closer to the combined sub-space of the cat and the dog, than to other non-animal classes. Such desired arrangement of the sub-spaces could be naturally attained through employing the word-vector similarities of the tags during training. 


In this work, we propose a deep neural network, complete with a learning algorithm, for weakly supervised learning of a semantic hashing model through using the word embeddings of the image tags. To the best our knowledge, this is the first work to use an end-to-end deep model to learn hash vectors using images and tags alone (without using labels). On the particular task of image hashing, our method appears to be the first work on using word embeddings of tags in a weakly supervised setting. We evaluate our approach and report systematic comparison with relevant state-of-the-art, and our approach is shown to outperform existing unsupervised or weakly supervised hashing methods for semantic image retrieval. 


\section{Related Work}

Much effort in the area of semantic image hashing has been directed towards utilizing supervised methodologies to learn the hash space. While there is some work in the area of unsupervised hashing, very little attempt was made in the area of weakly supervised hashing. Since the number of weakly supervised hashing techniques are very limited in number, we compare our model to both weakly supervised and unsupervised methods during evaluation. On similar lines, in this section, we give a brief overview of the related work from both the areas. 

The foremost image hashing algorithm called the Locality Sensitive Hashing \cite{charikar2002similarity} works on the principle of projecting the data on to random hyperplanes and computing each bit based on which half-space the sample falls into. This algorithm is data-independent and therefore the produced hash codes do not capture the structure in the data. Several variants (\cite{dasgupta2011fast}, \cite{kulis2012kernelized}, \cite{chakrabarti2015bayesian}) have been proposed, all producing hash codes irrespective of the distribution of the data. 

Another paradigm of image hashing is the data-dependent hashing methods. Traditionally data-dependent methods have been formulated as independent feature learning and hash coding stages. However, with the advent of deep learning and the huge amount of data available, literature has moved towards learning hash codes as single stage algorithms,  which take in image pixels as inputs and directly learn the hash codes. This can also be interpreted as an inbuilt feature learning technique that does not require human intervention. 

Approaches such as \cite{gong2013iterative}, \cite{weiss2009spectral}, \cite{wang2012semi} are some representative works of non-deep learning based unsupervised learning. \cite{gong2013iterative} tried to minimize the quantization error between the real-valued uncorrelated feature vector and the binary code by finding a rotation of the zero-centered data. \cite{weiss2009spectral} showed the analogy between the problem of finding the optimal hash space distribution and graph partitioning algorithm and attempted the problem using spectral ways. \cite{wang2012semi} attempted the problem of learning hash spaces in a semi-supervised way by back propagating the classification loss over a limited labeled data-set and an entropy based loss over the entire labelled and unlabelled data-set.

Representative deep-learning-based unsupervised hashing algorithms include \cite{erin2015deep}, \cite{lin2016learning}, \cite{do2016learning}. The work of \cite{erin2015deep}, though being deep-learning-based, is not an end-to-end framework that can take in raw images and produce the hashes. They used GIST features as inputs to the neural network and learned the hash codes by minimizing the quantization loss, maximum variance loss, and the independent bit loss. The key idea of \cite{lin2016learning} is to produce rotation invariant binary codes and showed that they achieve state-of-art performance on three different tasks namely, image matching, image retrieval and object recognition. The approach of \cite{do2016learning} learns hash codes as the outputs of the hidden layer of a binary  auto-encoder. This makes the learning problem NP-hard and they resort to an alternate optimization scheme to move towards the desired hash space. 

Another note-worthy mention in the area of uni-modal image hashing is \cite{cao2017deep}. They utilized the word embedding of labels as the supervision to learn an image hash space. While this appears similar to our work, they used vector representations of labels, rendering the work to fall under the category of supervised image hashing, whereas our work uses vector representations of raw tags.

A common characteristic among most deep learning and non-deep-learning based semantic hashing methods is that they rely only on the information from the images to learn the hash codes, often completely ignoring other associated metadata. Several works (\cite{jiang2016deep}, \cite{cao2016deep}, \cite{xu2017learning}) in the area of Cross Modal Hashing (CMH) attempted utilizing tag information along with image data to learn the hash space. However, as mentioned previously, they learn a common hash space for various modalities of input (image and tag in this case), which is different from what we intend to do. Among all the CMH methods, \cite{cao2016deep} is the closest approach to our work. \cite{cao2016deep} intends to align the visual space of images and the semantic space of sentences using language (\textit{word2vec}) and vision (CNN based) models. The main difference between their work and ours is that we attempt to use tag information which is much noisy than the actual English sentences they used in their work. Practically, such clean English sentences are as hard to obtain as the supervised label information. An extensive discussion on CMH and uni-modal hash learning can be found in \cite{wang2016comprehensive} and \cite{wang2017survey} respectively. 




Unlike CMH, weakly supervised hashing methods leverage only the image-tag information during training. \cite{guantag}, \cite{zhang2016discrete}, \cite{tang2017weakly} are some well-known works in this area. The authors of \cite{guantag} proposed a framework which consist of two stages, weakly supervised pre-training and fine-tuning using supervised labels. \cite{zhang2016discrete} used collaborative filtering with hashing in predicting the image-label associations, where the ground-truth labels are used to generate the label matrix. To our best knowledge, \cite{tang2017weakly} is the only prior approach that attempted truly weekly supervised hashing (i.e., without using label information). More specifically, they attempted to explore the discriminative information and the local geometric structure from the tags and images. They then formulated the hashing problem as an eigenvalue problem. Considering these facts, we only compare our approach to \cite{tang2017weakly} among the weekly-supervised methods.



In this work, we intend to build an end-to-end deep learning hashing model that does not require expensive labels in training but can still generate semantically meaningful hash codes. In the experiments section, we compare our model to the following unsupervised and weakly supervised image hashing approaches: \cite{gong2013iterative}, \cite{weiss2009spectral}, \cite{wang2012semi}, \cite{wang2012semi}, \cite{jin2014density}, \cite{heo2012spherical}, \cite{zhu2014sparse}, \cite{liu2014discrete}, \cite{do2016learning}, \cite{zhang2016discrete}, \cite{tang2017weakly}. Additionally, in-order to show the significance of the usage of tag embeddings, we developed a deep learning based baseline which intends to learn a semantic hash space using only the binary tag vectors. More details about our approach and the binary tag vector model are presented in the next section. 

\section{Proposed Approach}
\begin{figure*}[!t]
\includegraphics[width=1.0\textwidth]{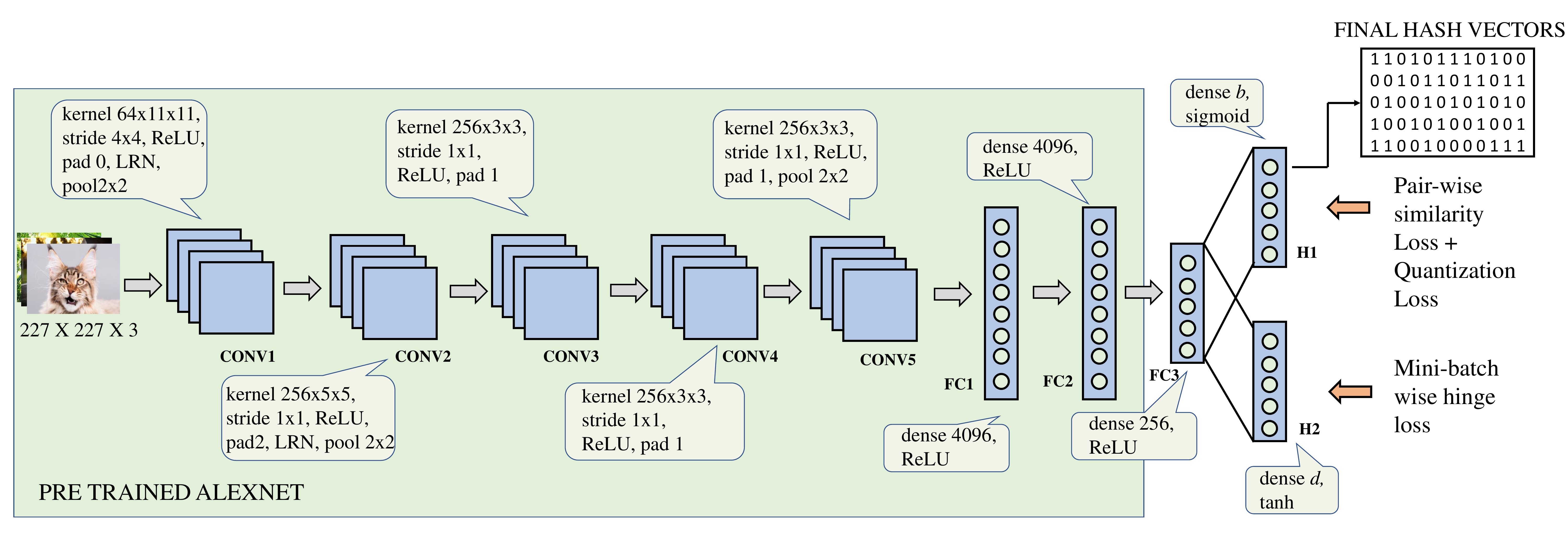} 
\caption{Network Architecture of the proposed model. The outer green box represents the pre-trained AlexNet model; The FC3, H1 and H2 layers are the newly appended layers. The final hash codes are extracted from the H1 layer. }
\label{fig:arch}
\end{figure*}

\subsection{Problem Formulation}
In this work we assume that the datasets have triplets of image-tags-labels ( $\textbf{\textit{x}}_i$, $T_i$, $\textbf{\textit{l}}_i$). Here, $\textbf{\textit{x}}_i$ represents the image feature vector for the $i^{th}$ sample, $T_i$ represents the corresponding tags set and $\textbf{\textit{l}}_i$ represents its binary label vector. In a generic scenario, each sample is associated with more than one tag and more than one semantic label. Therefore, the tags are represented as a set $T_i$ and the labels are represented as a binary vector $\textbf{\textit{l}}_i$. In the label vector, the value of an element is $1$ if the corresponding label is associated with that image and is $0$ otherwise. Our task is to find a function {\Large {$\Psi (\cdot) $}} that takes ($\textbf{\textit{x}}_i$, $T_i$) as inputs and produces a hash vector $\textbf{\textit{b}}_i$ as output. The hash space thus learnt should map semantically similar images, defined by the label vectors, to nearby hash codes and dissimilar ones to farther codes. While the labels assumed to be unavailable during the training phase, they are employed during the testing phase to measure the performance of the learnt model. 
\subsection{Tag Processing}
Let $\tau_i^j$ represent a tag in the tag set $T_i$,. where $j$ is the index of the tag in the set, i.e., $j \in [1, m]$ where $m$ is the total number of tags associated with the $i^{th}$ sample. We convert each tag  $\tau_i^j$ into a $d$-dimensional vector using the \textit{word2vec} language model \cite{mikolov2013efficient}. Thus for each tag $\tau_i^j$, we obtain a vector representation $\textbf{\textit{v}}_i^j$ which is the \textit{word2vec} representation of the tag word $\tau_i^j$. Since each image has multiple tags associated with it, we aggregate all the tag vectors into a single $d$-dimensional vector for a given image. In this work, we adopted basic functions like \textit{tf}(tag frequency), \textit{itf}(inverse tag frequency)  and \textit{mean} to compute the aggregated vector $\textbf{\textit{w}}_i$. In experiments, we will compare these aggregation techniques by their performance. 

The formulae used to compute $\textbf{\textit{w}}_i$ are given below. \\
\begin{equation}
\begin{split}
&mean:     \textbf{\textit{w}}_i = \frac{1}{m}\sum_{j=1}^{m} \textbf{\textit{v}}_i^j \\
&\textit{tf}:     \textbf{\textit{w}}_i = \frac{1}{m}\sum_{j=1}^{m} \frac{n(\tau_i^j)}{ N } \textbf{\textit{v}}_i^j \\
&itf:   \textbf{\textit{w}}_i = \frac{1}{m}\sum_{j=1}^{m} \log \frac{N}{n(\tau_i^j)} \textbf{\textit{v}}_i^j
\end{split}
\end{equation}
Here, $N$ represents the total number of tags in the database and $n(\tau_i^j)$ represents the number of images associated with the tag $\tau_i^j$. 

Thus we arrive at the \textit{image - tag vector} ($\textbf{\textit{x}}_i$, $\textbf{\textit{w}}_i$) pairs from the initial \textit{image - tag set} ($\textbf{\textit{x}}_i$, $T_i$) pairs. 
\subsection{Designing a Network for Hashing}

 We use the pre-trained AlexNet model as a key building block for our hashing model. The network takes 227X227X3 dimensional images as input and passes them through five convolutional layers and two fully connected layers, labelled as  CONV$i$ ($i$=1,...,5), FC1 and FC2. Until the FC2 layer, the architecture is identical to the AlexNet \cite{krizhevsky2012imagenet} architecture and the weights are initialized to the pre-trained ImageNet \cite{deng2009imagenet} weights. The FC2 layer produces a $4096$ dimensional vector, which is given as input to another fully connected layer FC3. FC3 outputs $256$ dimensional vector which is further fully connected to two layers H1 and H2 in a lateral fashion. The acronyms H1 and H2 represent the \textit{Head1} and \textit{Head2} respectively. The outputs of H1 and H2 are $b$ (number of bits in the hash code) and $d$ (dimensionality of the aggregated tag vector) dimensional vectors, which are then topped by \textit{sigmoid} and \textit{tanh} activations respectively. The overall model is shown in Figure 1. The new layers beyond the AlexNet layers are initialized with \textit{glorot normal} \cite{glorot2010understanding} weights. The VGG-19 network was also attempted, giving results similar to those of the AlexNet model but with much-increased training time. We therefore decided to train all our models using the Alexnet model.

The model is trained on three loss components back propagating from the two heads H1 and H2 into the network. More specifically, we back propagate pair-wise similarity loss and quantization loss from H1, and mini-batch wise hinge loss from H2. Thus we presume that the loss on H2 (hinge loss) forces the network to form feature spaces (especially at the later layers, H2 and FC3) that are in accordance with the semantic information contained in the aggregated tag vectors, $\textbf{\textit{w}}_i$. On the other hand, the pair-wise loss on H1 aligns the hash space such that semantically similar image pairs are close by and dissimilar pairs are farther. Thus the two main loss components augment each other and guide the network towards learning a semantically meaningful hash space. The third loss component, the quantization loss, forces the output of H1 to be close to $0$ or $1$. 

The pairwise Euclidean loss applied on H1 was first used for hashing in \cite{kulis2009learning} while the quantization loss was first used in \cite{gong2013iterative}. The hinge loss on the head H2 is a ranking loss first used by \cite{frome2013devise} to learn a semantically meaningful real valued image representation using word embeddings of classification labels. While the hinge loss component does not seem to serve a clear purpose in this network architecture, empirical results show that this component contributes significantly to the performance boost of our model. Also, \cite{frome2013devise} mentions that using such loss boosts the performance of their model instead of using a $L2$ component. They presume that this could be due to the fact that the problem of forming a semantically meaningful image representation space is a ranking problem in general and therefore such a ranking loss could be more relevant. On similar lines, we can argue that the current problem of learning image hashes is a ranking problem as well, and thus, such a hinge loss component could boost the performance of a retrieval system significantly.

During inference, only H1 is used to extract the features, which are then  quantized to obtain the hash code according to the following scheme: $\textbf{\textit{b}}_i = \frac{1}{2}(sgn(\textbf{\textit{h}}_i^{(1)} - 0.5\textbf{\textit{1}}) + 1)$. Here, $\textbf{\textit{h}}_i^{(1)}$ represents the real-valued feature vector obtained at the output of H1, $sgn$ represents a sign function that outputs $1$/$-1$ based on if the input to the sign function is positive or negative and lastly, $\textbf{\textit{1}}$ represents a vector of ones of length $b$. Thus, we obtain binary codes which have a value of $1$/$0$ from a raw train/test images. 
\subsection{Designing the Loss Functions}
\textbf{Pair-wise Similarity Loss:}
Most state-of-art supervised learning methods assume binary similarity between two images, i.e., two images can be either similar($1$) or dissimilar($0$) depending on if they share a common label or not. However, in the current weakly supervised learning context, we intend to use cosine similarity between the aggregated tag vectors as the ground truth similarity. Since cosine similarity is real-valued and can take values between -1 and 1, the ground truth similarity in our case is not binary valued, i.e., we can deem an image pair to be less similar or more similar, instead of absolutely declaring it to be similar or dissimilar. We only consider this notion of ground truth similarity during training and stick with the $0$/$1$ similarity during evaluation.

We formulate the pair-wise similarity loss function as follows. For any image pair ($\textbf{\textit{x}}_i$, $\textbf{\textit{x}}_j$), the loss function should push the corresponding hashes closer if the cosine distance between them is smaller and vice-versa. The equation of this loss function is given below, 
\begin{equation}
\begin{split}
     L_1 &= \sum_{i=1}^k \sum_{j=1}^k [\frac{1}{b}(\textbf{\textit{h}}_i^{(1)} - \textbf{\textit{h}}_j^{(1)})^T \cdot (\textbf{\textit{h}}_i^{(1)} - \textbf{\textit{h}}_j^{(1)})\\ 
     &- \frac{1}{2}(1.0 - \frac{\textbf{\textit{w}}_i^T \cdot \textbf{\textit{w}}_j}{\|{\textbf{\textit{w}}_i}\| \|{\textbf{\textit{w}}_j}\|})]^2
\end{split}
\end{equation}
where $k$ is the mini batch size and the two summations signify computing pairwise losses across all possible pairs. The vectors $\textbf{\textit{h}}_i^{(1)}$ and $\textbf{\textit{h}}_j^{(1)}$ represent the output vectors of H1 for sample $\textbf{\textit{x}}_i$ and $\textbf{\textit{x}}_j$ respectively.
A lower value of $L_1$ is obtained when a high value of $1.0 - \frac{\textbf{\textit{w}}_i^T \cdot \textbf{\textit{w}}_j}{\|{\textbf{\textit{w}}_i}\| \|{\textbf{\textit{w}}_j}\|}$ results in a high value of $(\textbf{\textit{h}}_i^{(1)} - \textbf{\textit{h}}_j^{(1)})^T \cdot (\textbf{\textit{h}}_i^{(1)} - \textbf{\textit{h}}_j^{(1)})$ and vice-versa. Higher value of $1.0 - \frac{\textbf{\textit{w}}_i^T \cdot \textbf{\textit{w}}_j}{\|{\textbf{\textit{w}}_i}\| \|{\textbf{\textit{w}}_j}\|}$ is obtained when the samples are dissimilar, thus the hash codes should be pushed apart. Similarly, lower value of this term is obtained when the samples are similar and therefore the hash codes should be pushed closer. 

\textbf{Mini-batch-wise Hinge Loss:} In addition to the pairwise similarity loss, we also intend to back-propagate a loss that forms a semantic embedding space at the output of H2. Such a loss function adjusts the feature spaces of not only the H2 layer but also some of the previous layers (FC3, FC2), thus transmitting the semantic information from the tags back into the network. As H1 is connected to the output of FC3, the semantic information contained in FC3 will aid in learning the hashes at the output of H2, thus enhancing the model's performance. 
To this end, we define the following loss, 
\begin{equation}
    L_2 = \sum_ n \sum_{j\ne n} max[0, margin + \textbf{\textit{w}}_j \cdot \textbf{\textit{h}}_n^{(2)} - \textbf{\textit{w}}_n \cdot \textbf{\textit{h}}_n^{(2)}]
\end{equation}
where $\textbf{\textit{h}}_n^{(2)}$ represents the output of the head H2 for the $n^{th}$ sample in the mini-batch. The loss $L_2$ is $0$ only when the quantity $\textbf{\textit{w}}_n \cdot \textbf{\textit{h}}_n^{(2)}$ is more than $margin + \textbf{\textit{w}}_j \cdot \textbf{\textit{h}}_n^{(2)}$. That is, the value of the loss is zero only when the prediction of head $H2$ for the $n^{th}$ sample is closer to the ground truth aggregated tag vector $\textbf{\textit{w}}_n$ than to any other ground truth tag vector $\textbf{\textit{w}}_j$ by a margin $margin$. A similar idea was previously considered in \cite{hu2016video2vec}, where the goal was to semantically embed videos onto a space using the \textit{word2vec} representation of the video labels. As such, their approach is supervised (i.e., assuming the label information).

\textbf{Quantization Loss:} We further impose the quantization loss on the H1 output to force the outputs to be close to $0$ or $1$, as follows, 
\begin{equation}
    L_3 = - \sum_{i=1}^k \frac{1}{b} (\textbf{\textit{h}}_n^{(1)} - 0.5 \textbf{\textit{1}})^T \cdot (\textbf{\textit{h}}_n^{(1)} - 0.5 \textbf{\textit{1}})
\end{equation}
This function penalizes the network if the output of a neuron is close to 0.5. 

During training, we weigh the three loss components $L_1$, $L_2$ and $L_3$ by factors $\lambda_1$, $\lambda_2$ and $\lambda_3$ respectively. Therefore the resultant loss that will be back-propagated is: $L = \lambda_1 L_1 + \lambda_2 L_2 + \lambda_3 L_3$
\subsection{The binary tag-vector model}
In addition to comparing our method with several state-of-art models, we built another deep model, that uses the binary tag vectors for supervision, unlike the \textit{word2vec} tag embeddings we used in WDHT. We call this model the \textit{binary tag-vector model} in the rest of the text. To accommodate this, we make slight modifications to our model. Firstly, we suppose that two images are similar if both of them share at least one tag. Such kind of formulation has been used in various supervised learning methods where they consider two images to be similar if both of them share at least one label. Since our problem setting is weakly supervised, we use tag vectors instead of label vectors. Tag vectors are binary vectors whose length is equal to the total number of tags in the data-set and will have a value of $1$ if the tag is associated with the image and $0$ otherwise. 

Regarding the network architecture, only the head H1 is kept and H2 is completely removed. We do this owing to the fact that the real-valued vectors (like aggregated tag vectors in the above scenario) are not available in this case, to regress the outputs to. Additionally, in the previous case, the loss applied on H1, i.e., the $L1$ component has a real-valued ground truth similarity, unlike the current scenario. Therefore, we use a different loss component (contrastive loss) to accommodate the binary valued ground truth similarity labels. The equation of the loss is as follows, 
\begin{equation}
    \begin{split}
        L_4 &= \sum_{i=1}^k \sum_{j=1}^k S*(1-\beta)*D +\\ &(1-S)*\beta * (max(0, margin - D))^2\\
        &\mathrm{where}\quad  D = \frac{1}{b}{(\textbf{\textit{h}}_i^{(1)} - \textbf{\textit{h}}_j^{(1)})}^T \cdot (\textbf{\textit{h}}_i^{(1)} - \textbf{\textit{h}}_j^{(1)})
    \end{split}
\end{equation}
Here, $margin$ represents the margin associated with the hinge loss component of the contrastive loss, $S$ represents the ground truth similarity label, and $\beta$ represents the fraction of similar sample pairs present in the mini batch. Weighing the loss sub-components by $\beta$ and $1-\beta$ respectively are important due to the fact that in any mini-batch only a small fraction of the image pairs will have at least one tag in common, thus making the dataset highly imbalanced. We therefore incorporate $\beta$ weight factor in the loss.

Thus the final loss for the binary tag-vector model becomes: $L =  \lambda_3 L_3 + \lambda_4 L_4$

\section{Experiments and Results}
\begin{table}[t]
\centering
\scalebox{0.9}{\begin{tabular}{|c|c|c|c|c|}
        \hline
        Method & 12 bits & 24 bits & 32 bits & 48 bits \\
        \hline
        \hline
        \textit{  itf  } & 0.6124 & 0.6323 & 0.6531 & 0.6644 \\
        \hline
        \textit{ tf} 
        &0.6394 
        &\textbf{0.6836}  &0.6881  
        &\textbf{0.6835} \\
        \hline
        \textit{ mean} & \textbf{0.6709} & 0.6805 & \textbf{0.6955} & 0.6621\\
        \hline
\end{tabular}}
\caption{Comparing the mAP of the model with the  \textit{itf}, \textit{tf} or \textit{mean} aggregation functions for the NUS-WIDE dataset. }\label{tab:cap}
\end{table}

\begin{table*}[t]
\centering

\scalebox{0.95}{\begin{tabular}{|c|c|c|c|c|c|c|c|c|c|}
        \hline
        \textbf{Algorithm} & \multicolumn{4}{|c|}{\textbf{NUS-WIDE}} & \multicolumn{4}{|c|}{\textbf{MIRFLICKR-25K}} \\
        \hline
        { } & 12bits & 24bits & 32bits & 48bits & 12bits & 24bits & 32bits & 48bits\\
        \hline
        \hline 
        ITQ  \cite{gong2013iterative}\textit{(non-deep)}&0.5295&0.5227&0.4932&0.5275&0.6418&0.655&0.6253&0.6504\\
        \hline
        PCAH \cite{wang2012semi}\textit{(non-deep)}&0.4566&0.4209&0.4016&0.3971&0.6098&0.6033&0.6085&0.6169\\
        \hline
        LSH \cite{charikar2002similarity}\textit{(non-deep)}&0.3308&0.3682&0.3726&0.3918&0.5708&0.5885&0.5843&0.6015\\
        \hline
        DSH \cite{jin2014density}\textit{(non-deep)}&0.5065&0.5118&0.4902&0.4807&0.6561&0.6593&0.644&0.6422\\
        \hline
        SpH \cite{heo2012spherical}\textit{(non-deep)}&0.3829&0.3959&0.3907&0.3947&0.586&0.5785&0.5789&0.5789\\
        \hline
        SH \cite{weiss2009spectral}\textit{(non-deep)}&0.4503&0.4029&0.4006&0.3731&0.6251&0.6157&0.6044&0.596\\
        \hline
        AGH \cite{liu2014discrete}\textit{(non-deep)}&0.535&0.5226&0.497&0.4791&0.6378&0.6484&0.6473&0.6346\\
        \hline
        DH \cite{erin2015deep}\textit{(deep)}&0.4036&0.3974&0.3932&0.4014&0.5833&0.5945&0.5932&0.5942\\
        \hline
        UH-BDNN \cite{do2016learning}\textit{(deep)}&0.4982&0.4996&0.4823&0.4853&0.6324&0.6279&0.6274&0.6258\\
        \hline
        DeepBit \cite{lin2016learning}\textit{(deep)}&0.4225&0.4247&0.4359&0.431&0.5974&0.6032&0.6077&0.6115\\
        \hline
        Binary Tag Vector\textit{(deep)}&0.4809&0.475&0.4793&0.4702&0.6064&0.6087&0.6077&0.6098\\
        \hline
        Proposed(WDHT)\textit{(deep)}&\textbf{0.6258}&\textbf{0.6397}&\textbf{0.6606}&\textbf{0.647}&\textbf{0.687}&\textbf{0.695}&\textbf{0.6667}&\textbf{0.6621}\\
        \hline
        \hline
        WMH*\textit{(non-deep)}&0.299&0.306&0.307&0.309&0.585&0.590&0.582&0.573\\
        \hline
        Proposed(WDHT*)\textit{(deep)}&0.4910&0.4916&0.4835&0.485&0.626&0.6355&0.6326&0.6308\\
        \hline
\end{tabular}}
\caption{MAP values of NUS-WIDE and MIR-FLICKR25k data-sets computed using the top 50,000 retrieved images.} \label{tab:cap}
\end{table*}

\begin{table*}[t]
\centering

\scalebox{0.98}{\begin{tabular}{|c|c|c|c|c|c|c|c|c|c|}
        \hline
        \textbf{Algorithm} & \multicolumn{4}{|c|}{\textbf{NUS-WIDE}} & \multicolumn{4}{|c|}{\textbf{MIRFLICKR-25K}} \\
        \hline
        { }& 12bits & 24bits & 32bits & 48bits & 12bits & 24bits & 32bits & 48bits\\
        \hline
        \hline 
        ITQ \cite{gong2013iterative}\textit{(non-deep)}&0.6329&0.6299&0.594&0.6478&0.6908&0.7064&0.6684&0.6996 \\
        \hline
        PCAH \cite{wang2012semi} \textit{(non-deep)}&0.5766&0.5046&0.49&0.4904&0.643&0.6306&0.6372&0.6516 \\
        \hline
        LSH \cite{charikar2002similarity}\textit{(non-deep)}&0.3501&0.4093&0.4169&0.4546&0.5736&0.6049&0.5954&0.6239 \\
        \hline
        DSH \cite{jin2014density}\textit{(non-deep)}&0.5919&0.5982&0.5713&0.5791&0.6955&0.7071&0.6834&0.6603 \\
        \hline
        SpH \cite{heo2012spherical}\textit{(non-deep)}&0.4645&0.4645&0.4465&0.4472&0.5966&0.5811&0.5828&0.579 \\
        \hline
        SH \cite{weiss2009spectral}\textit{(non-deep)}&0.5623&0.5033&0.4896&0.4533&0.6605&0.6405&0.6291&0.6213 \\
        \hline
        AGH \cite{liu2014discrete}\textit{(non-deep)}&0.6551&0.6459&0.6274&0.6225&0.6862&0.7005&0.6998&0.6853 \\
        \hline
        DH \cite{erin2015deep}\textit{(deep)}&0.4733&0.4601&0.462&0.4763&0.6033&0.6195&0.6135&0.618 \\
        \hline
        UH-BDNN \cite{do2016learning}\textit{(deep)}&0.5923&0.5915&0.5902&0.6097&0.6654&0.6684&0.6672&0.6699 \\
        \hline
        DeepBit \cite{lin2016learning}\textit{(deep)}&0.5463&0.5548&0.5624&0.561&0.589&0.6027&0.609&0.6086 \\
        \hline
        Binary Tag Vector\textit{(deep)}&0.6202&0.627&0.6247&0.6249&0.6365&0.6326&0.6373&0.6352 \\
        \hline
        Proposed(WDHT)\textit{(deep)}&\textbf{0.6709}&\textbf{0.6805}&\textbf{0.6955}&\textbf{0.676}&\textbf{0.7346}&\textbf{0.743}&\textbf{0.7034}&\textbf{0.7054} \\
        \hline
\end{tabular}}
\caption{MAP values of NUS-WIDE and MIR-FLICKR25k data-sets computed using the top 5,000 retrieved images} \label{tab:cap}
\end{table*}
\begin{figure*}[!t]
\centering
\includegraphics[width=0.8\textwidth]{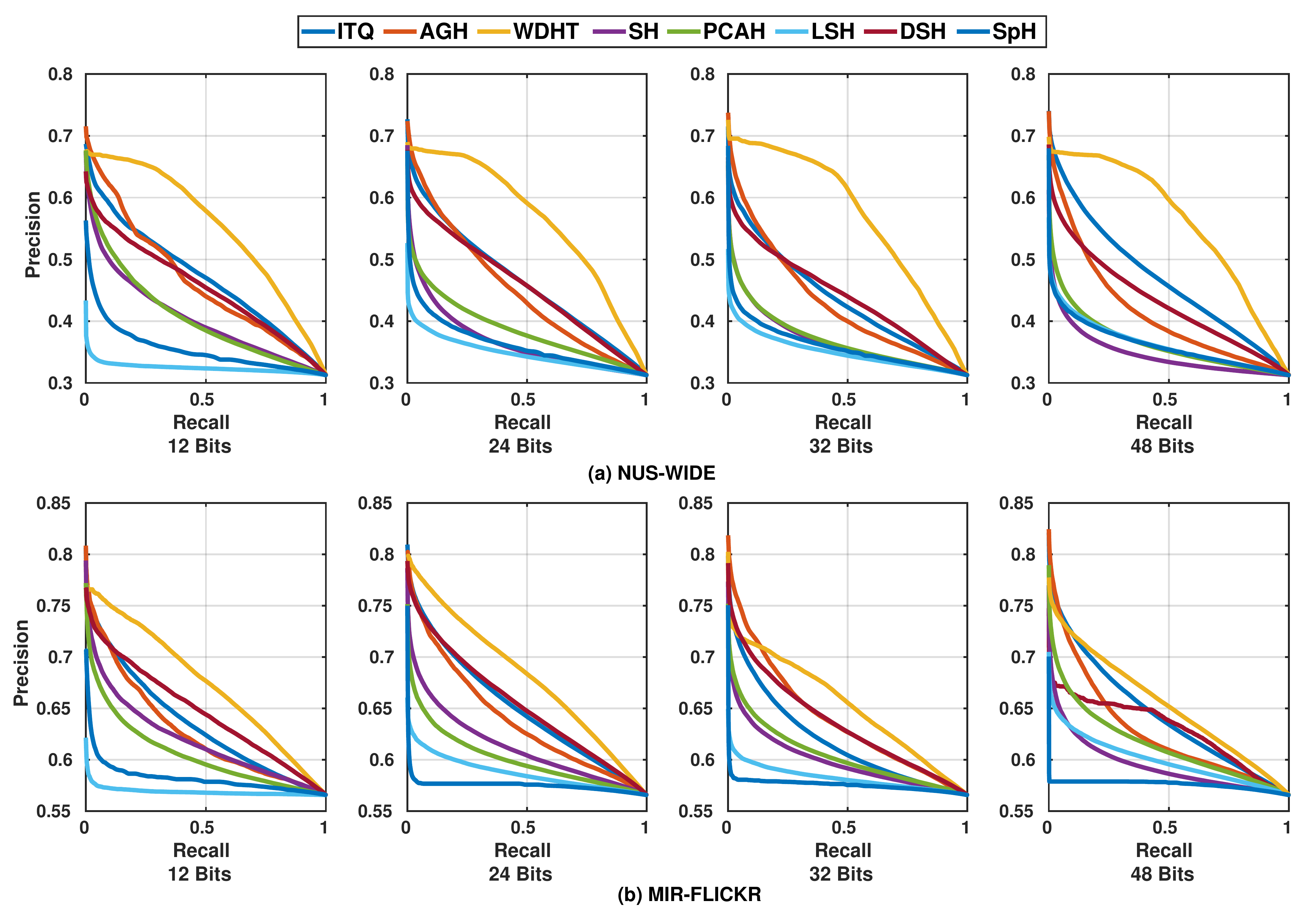} 
\caption{Precision Recall curves for NUS-WIDE and MIR-FLICKR datasets.}
\label{fig:arch}
\end{figure*}
\subsection{Datasets}
{\bf{NUS-WIDE}} This is a Web image dataset with 269,648 images collected from Flickr. Each image is associated with a set of tags. \cite{chua2009nus} presents that there is a total of 425,059 tags associated with the 269k images. Further, the authors of \cite{chua2009nus} conducted manual annotation of these images to a predefined set of 81 labels. For our experiments, we used only the images that are associated with at least one of the 21 most frequent labels. Thus we formed a training set of 100,000 images and a testing set of 2,000 images. We used the whole training set as the database and the testing set as the query set during evaluation. {\bf{MIR-FLICKR25K}} This is a comparatively smaller dataset with 25,000 images collected from Flickr and contains 1386 tags associated with them. \cite{huiskes2008mir} manually associated the images with 38 semantic categories. For our experiments, we used the images which are associated with at least one of the 38 categories. Thus we used a total of 16,000 images for training and 2,000 for testing. For both the data-sets, we randomly picked the testing set without considering the labels of the images.

\subsection{Training}
We trained our model using mini-batch gradient descent with a learning rate of 0.001 for the last three layers (FC3, H1, and H2) and a learning rate of 0.0001 for the pre-trained layers (CONV1 - FC2). We also used the momentum term with the rate of momentum equal to 0.9. The weighing factors for the losses, $\lambda_1$, $\lambda_2$, $\lambda_3$ and $\lambda_4$, are set to 1.0, 10.0, 1.0 and 1.0 respectively for all the experiments, which were determined by performing a grid search over the hyper-parameter space. The \textit{word2vec} model that we used was trained on 1 billion words from the Wikipedia documents and outputs a 300-dimensional vector for a given word. Therefore the number of output neurons on H2 is set to 300. 

\subsection{Performance Evaluation}

We evaluated the learned hash codes for the task of semantic image retrieval. We used the mean-Average-Precision (mAP) metric to compare our model's performance to the existing methods. We used the same protocol used by \cite{li2015feature}, \cite{wang2016deep}, \cite{lai2015simultaneous} and several others to compute the mAP values. The results are compared against eleven state-of-art approaches ITQ, PCAH, LSH, DSH, SpH, SH, AGH, DH, UH-BDNN, DeepBit and WMH. All the methods, except WMH are run using the code provided by the authors and for the suggested hyper-parameter settings. As most of the works presented here are based on the pre-determined feature vectors, we extracted the 4096-dimensional vectors from the AlexNet model (i.e the output of FC2) and used them as input to these methods. For WMH we directly quote the results from the original paper (the code is not publicly available). For a fair comparison, we run our model with the same experimental setting as WMH and report the results. We first filtered the images and tags in WMH's standard, then performed another round of experiments using only 5,000 training images and 1,000 query images for the two datasets.

Firstly, to finalize the tag aggregation scheme, we compared the performance of our model using the \textit{itf}, \textit{tf} and \textit{mean} functions for aggregation on NUS-WIDE data-set. We noticed that \textit{mean} worked slightly better than the \textit{idf} and \textit{tf} as can be seen from Table 2. Further, we performed a variance analysis on the word vectors of tags associated with each image. More specifically, we computed the variance of the tag vectors for each image and then analyzed the histogram of the variances for all images. It was found that a majority of the variances falls below 8. Note that the maximum distance between any two word vectors in this space can be $2\sqrt{300}$ (the range of each dimension of the tag vector is $[-1, 1]$ and the space is 300-dimensional). This appears to suggest that for most of the images, their tag vectors do not spread out too much, which might explain that the simple \textit{mean} aggregation function is working reasonably well. 

Further, we computed the mAP for two different settings, one using the top 50,000 retrieved images and another using the top 5,000 retrieved images for the unsupervised approaches and report the results in Table 3 and Table 4 respectively. The first seven methods presented here are non-deep-learning methods while the last three are deep-learning-based. Additionally, DH \cite{erin2015deep} and UH-BDNN \cite{do2016learning}, even though being deep-learning-based, depend on the hand-crafted features. DeepBit \cite{lin2016learning} is the only work that takes a raw image as input and produces a binary code, but its performance is inferior to most other methods. In contrast, our approach (WDHT) is an end-to-end framework and performed superior than all the state-of-art methods on both datasets.
 
The non-deep-learning based approaches ITQ \cite{gong2013iterative} and AGH \cite{liu2014discrete} seem to stand in the second and the third places in terms of the mAP values in the experiments. These methods are performing superior to the existing deep learning based methods (\cite{erin2015deep}, \cite{do2016learning}, \cite{lin2016learning}) as well. On the other hand, the weakly supervised approach WMH seemed to perform quite inferior as compared to WDHT with the new experiment setting. The results are presented as the bottom 2 rows of Table 3. 

For further analysis, we plotted the precision-recall curves in Figure 2. These curves are computed taking into consideration all the retrieved samples from the database for a given query image. More specifically, we computed the average precision for various values of recall (1000 discrete values of recall) for all query images. The big performance gain of our approach on the NUS-WIDE data-set can be noticed from these graphs as well. 


\begin{figure}
\centering

\begin{minipage}{.4\textwidth}
  \centering
  \includegraphics[width=0.85\linewidth]{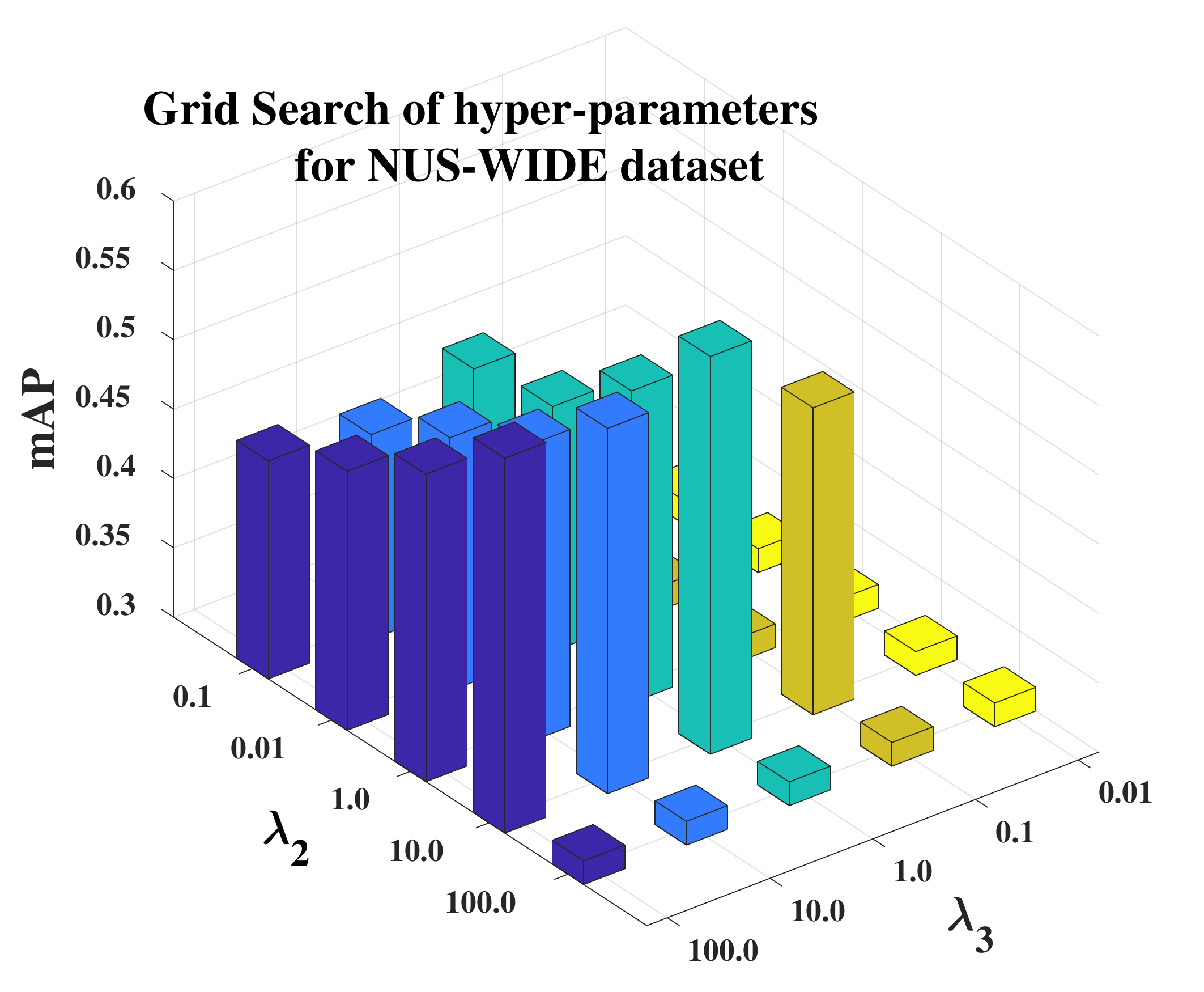}
  \captionof{figure}{MAP values obtained for various hyper-parameter settings for the NUS-WIDE dataset}
  \label{fig:test1}
\end{minipage}%
\end{figure}

The presence of three loss components in the objective functions triggers the obvious question of combining them in the right proportions. To analyze this, we fix the value of $\lambda_1$ to 1.0 and change the values of $\lambda_2$ and $\lambda_3$ between $0.01$ and $100.0$.  We performed a grid search over this range and chose the best hyper-parameters for our final model. Specifically, we set three values to the ones that gave maximum mAP value over a validation set during the grid search. For each setting of the hyper-parameter values, we only used 10,000 training sample due to the high training time of these experiments. A bar plot of the validation mAPs of NUS-WIDE dataset for various values of $\lambda_2$ and $\lambda_3$ is given in Figure 3. It can be noticed that higher values of $\lambda_2$ and lower values of $\lambda_3$ gave significantly better mAP as compared to other combinations. A similar behaviour was noticed on the MIR-FLICKR dataset as well. This is in accordance with the rationale presented in Section 3.3 that a ranking loss is better at forming semantically meaningful spaces as compared to Euclidean loss components (\cite{frome2013devise}). While this rationale is yet to be validated mathematically, our results suggest this seems to be the case empirically.


\section{Conclusion}
In this paper, we attempted the problem of weakly supervised deep image hashing using tag embedding. Our method is an end-to-end framework that takes raw images and tags as inputs and produces hash codes. Therefore, our model is applicable to Web images where such information is abundant. Through extensive experiments with comparison with existing state-of-the-art, we demonstrated that the proposed approach was able to deliver significant performance boost when evaluated on two well-known and widely-tested datasets. Future work includes possible better aggregation schemes in the \textit{word2vec} space that may lead to improved performance. 

\clearpage
{\small
\bibliographystyle{unsrt}
\bibliography{egbib}
}

\end{document}